\documentclass[10pt,twocolumn,letterpaper]{article}

\usepackage[pagenumbers]{cvpr} 

\usepackage[utf8]{inputenc} 
\usepackage{booktabs}       
\usepackage{amsfonts}       
\usepackage{nicefrac}       
\usepackage{microtype}      
\usepackage{xcolor}         
\usepackage{soul}
\usepackage{makecell}
\usepackage{multirow}
\usepackage{threeparttable}
\usepackage{arydshln}
\usepackage{lipsum}

\usepackage[ruled,noline,noend]{algorithm2e}

\usepackage{amsmath}
\usepackage{listings}
\usepackage{color}
\usepackage{courier}

\usepackage{amsmath}
\usepackage{natbib}
\usepackage{float}
\usepackage{subcaption}
\usepackage{tabularx}
\usepackage[most]{tcolorbox}

\usepackage{bbding}

\definecolor{codegray}{gray}{0.95}
\usepackage{balance}

\definecolor{cvprblue}{rgb}{0.21,0.49,0.74}
\usepackage[pagebackref,breaklinks,colorlinks,allcolors=cvprblue]{hyperref}

\title{CUDA-LLM: LLMs Can Write Efficient CUDA Kernels}

\author{Wentao Chen\quad
Jiace Zhu\quad
Qi Fan\quad
Yehan Ma\quad
An Zou\thanks{Corresponding author.}\\
Shanghai Jiao Tong University\\
{\tt\small \{wentaochen, zhujiace, fanqi666, yehanma, an.zou\}@sjtu.edu.cn} \\
}

\begin{document}
\maketitle

\begin{abstract}
Large Language Models (LLMs) have demonstrated strong capabilities in general-purpose code generation. However, generating the code which is deeply hardware-specific, architecture-aware, and performance-critical, especially for massively parallel GPUs, remains a complex challenge. In this work, we explore the use of LLMs for the automated generation and optimization of CUDA programs, with the goal of producing high-performance GPU kernels that fully exploit the underlying hardware.
To address this challenge, we propose a novel framework called \textbf{Feature Search and Reinforcement (FSR)}. 
FSR jointly optimizes compilation and functional correctness, as well as the runtime performance, which are validated through extensive and diverse test cases, and measured by actual kernel execution latency on the target GPU, respectively.
This approach enables LLMs not only to generate syntactically and semantically correct CUDA code but also to iteratively refine it for efficiency, tailored to the characteristics of the GPU architecture.
We evaluate FSR on representative CUDA kernels, covering AI workloads and computational intensive algorithms. Our results show that LLMs augmented with FSR consistently guarantee correctness rates. Meanwhile, the automatically generated kernels can outperform general human-written code by a factor of up to 179$\times$ in execution speeds.
These findings highlight the potential of combining LLMs with performance reinforcement to automate GPU programming for hardware-specific, architecture-sensitive, and performance-critical applications. 
\end{abstract}
\section{Introduction}
\label{sec:intro}
Large Language Models (LLMs) have shown increasing promise in automating software development tasks, including generating code snippets, completing function bodies, and offering basic debugging assistance \cite{anthropic2024claude3,aws2023codewhisperer, roziere2023code,2024DeepSeek}. However, their effectiveness remains limited when the task shifts from general-purpose programming to generating code that is deeply \textbf{hardware-specific}, \textbf{architecture-aware}, and \textbf{performance-critical}, particularly when involving massively parallel computing on Graphics Processing Units (GPUs) \cite{ouyang2025kernelbench}.

GPU programming, especially in CUDA, involves much more than writing syntactically correct code \cite{2012CUDA}. Developers must carefully manage thousands of concurrent threads across thread blocks and warps, optimize memory access patterns, and balance resource usage — all while targeting a specific hardware configuration. This process is inherently hardware-specific, which means it requires detailed knowledge of the specific hardware features of certain GPUs, such as memory sizes, the number and types of hardware units, and resource constraints. It is also {architecture-aware}, requiring an understanding of the execution model of GPUs, which involves how threads are grouped, scheduled, and executed to avoid bottlenecks like warp divergence or uncoalesced memory access. Most importantly, GPU programming is \textbf{performance-critical}, where even small inefficiencies in kernel design or resource usage can cause significant slowdowns at scale.

These requirements pose a significant challenge for LLMs. While they may generate code that appears correct at a high level, they lack the architectural understanding to write and optimize low-level CUDA kernels effectively. The core question we explore is whether LLMs can be enhanced, not just to generate working CUDA code but to produce implementations that are tuned for real-world performance on actual hardware.
In this work, we introduce \textbf{Feature Search and Reinforcement (FSR)}, a framework that augments LLMs with performance optimization capabilities. FSR combines feature search and reinforcement to jointly target two critical objectives: (1) \textbf{functional correctness}, verified through test cases, and (2) \textbf{runtime performance}, measured by empirical execution latency on target GPUs.
We apply FSR to a variety of CUDA workloads, and our results show that it improves both correctness and performance. This allows LLMs to generate CUDA code that meets the demands of real-world deployment. By tightly integrating correctness checking with performance optimization, FSR moves LLMs closer to being practical code generation tools for hardware-specific, architecture-aware, and performance-sensitive applications.

\section{CUDA Kernels}
\label{sec:background_related}
CUDA (Compute Unified Device Architecture) \cite{2012CUDA,sanders2010cuda} is a parallel computing platform and programming model developed by NVIDIA that allows developers to tap into the massive parallelism of NVIDIA GPUs for general-purpose computing, often referred to as GPGPU (General-Purpose computing on Graphics Processing Units). While GPUs were originally designed for graphics rendering, CUDA extends the C/C++ language to allow developers to write code that executes directly on the GPU. This enables dramatic acceleration of compute-intensive workloads across domains such as scientific simulation, deep learning, and real-time image processing.

A typical CUDA application follows a \textbf{heterogeneous computing model}, where the CPU (host) manages control flow, memory allocation, and data transfer, while the GPU (device) handles data-parallel computation. The distinction between host and device code is fundamental to CUDA design. Functions that run on the GPU are known as \textbf{kernels}, marked with the $\verb|__global__|$ qualifier and launched from the host using the special $\verb|<<<gridSize, blockSize>>>|$ syntax. Kernels execute in parallel across thousands of lightweight GPU threads, which are organized hierarchically into thread blocks and grids.

The following example illustrates a complete CUDA program that performs element-wise addition of two vectors \cite{nvidia2025cudasamples}. Each GPU thread is responsible for computing one element of the output vector by adding the corresponding elements from the input vectors \texttt{a} and \texttt{b}.

\begin{lstlisting}[caption={Complete CUDA program for vector addition.}]
#include <iostream>
#define N 512 // Define the length of the vectors

// __global__ indicates this is a kernel function to be executed on the GPU
__global__ void vectorAdd(float *a, float *b, float *c) {
    int i = threadIdx.x + blockDim.x * blockIdx.x;
    if (i < N) {
        c[i] = a[i] + b[i];
    }
}

int main() {
    float *h_a, *h_b, *h_c;
    float *d_a, *d_b, *d_c;
    size_t size = N * sizeof(float);

    // Allocate and initialize host memory
    h_a = (float*)malloc(size);
    h_b = (float*)malloc(size);
    h_c = (float*)malloc(size);
    for (int i = 0; i < N; i++) {
        h_a[i] = i;
        h_b[i] = i * 2.0f;
    }

    // Allocate device memory
    cudaMalloc(&d_a, size);
    cudaMalloc(&d_b, size);
    cudaMalloc(&d_c, size);

    // Copy inputs to device
    cudaMemcpy(d_a, h_a, size, cudaMemcpyHostToDevice);
    cudaMemcpy(d_b, h_b, size, cudaMemcpyHostToDevice);

    // Launch kernel
    vectorAdd<<<(N + 255) / 256, 256>>>(d_a, d_b, d_c);

    // Copy result back to host
    cudaMemcpy(h_c, d_c, size, cudaMemcpyDeviceToHost);

    // Display results
    for (int i = 0; i < 10; i++) {
        std::cout << h_a[i] << " + " << h_b[i] << " = " << h_c[i] << std::endl;
    }

    // Clean up
    free(h_a);
    free(h_b);
    free(h_c);
    cudaFree(d_a);
    cudaFree(d_b);
    cudaFree(d_c);

    return 0;
}
\end{lstlisting}

This example demonstrates three core components of a CUDA program:

\begin{itemize}
\item \textbf{Host Code}: Manages initialization, memory allocation, kernel launch, and post-processing. This is written in conventional C/C++ and runs on the CPU.
\item \textbf{Memory Management}: Because host and device use separate memory spaces, developers must explicitly allocate memory with \texttt{cudaMalloc()}, and transfer data between host and device using \texttt{cudaMemcpy()}.
\item \textbf{Device (GPU) Kernel}: Implements the parallel logic. The \texttt{vectorAdd} function uses built-in CUDA thread identifiers like \texttt{threadIdx}, \texttt{blockIdx}, and \texttt{blockDim} to distribute work across thousands of threads.
\end{itemize}

\begin{figure*}[t]
    \centering
    \includegraphics[width=0.7\linewidth]{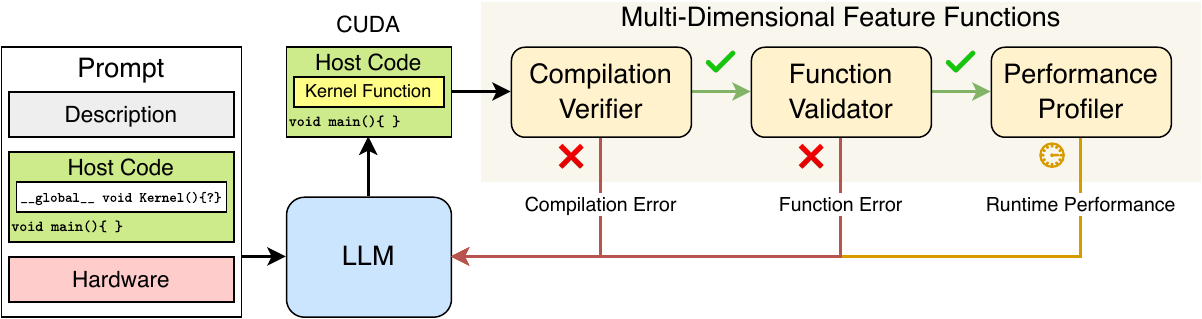}
    \caption{Overview of Feature Search and Reinforcement (FSR) framework.}
    \label{fig:overview}
\end{figure*}

While CUDA offers fine-grained control and exceptional performance for parallel workloads, writing CUDA code, especially device kernels, introduces challenges far beyond those encountered in Python or even traditional C development. Python is known for its high-level abstractions and ease of use, often relying on libraries like NumPy or PyTorch \cite{paszke2019pytorch} to internally optimize performance with C or CUDA backends. C, though lower-level, follows a sequential or moderately parallel model that does not require deep architectural knowledge of the underlying hardware.

Developing CUDA kernels requires a deep understanding of GPU execution models, thread hierarchies, and memory types (global, shared, local). Programmers must manage correct indexing, boundary checks, and synchronization to avoid race conditions and control-flow divergence. At the same time, achieving high performance demands careful optimizations, such as leveraging shared memory, maximizing memory coalescing, minimizing register spills, and tuning thread, and warp occupancy. This hardware-conscious nature makes CUDA development inherently complex and error-prone. Even small mistakes in thread coordination or memory access can lead to subtle bugs or significant performance loss.

In recent work, Ouyang \emph{et al.} \cite{ouyang2025kernelbench} propose the question: Can LLMs write efficient GPU kernels? Follow-up studies \cite{lange2025ai} have further investigated the translation of PyTorch models into CUDA kernels. Our work demonstrates that \textbf{LLMs Can Write Efficient CUDA Kernels} directly from natural language.

\section{Multi-Dimensional Feature Search and Reinforcement Framework}
\label{sec:frame}

\begin{figure*}[t]
    \centering
    \includegraphics[width=0.9\linewidth]{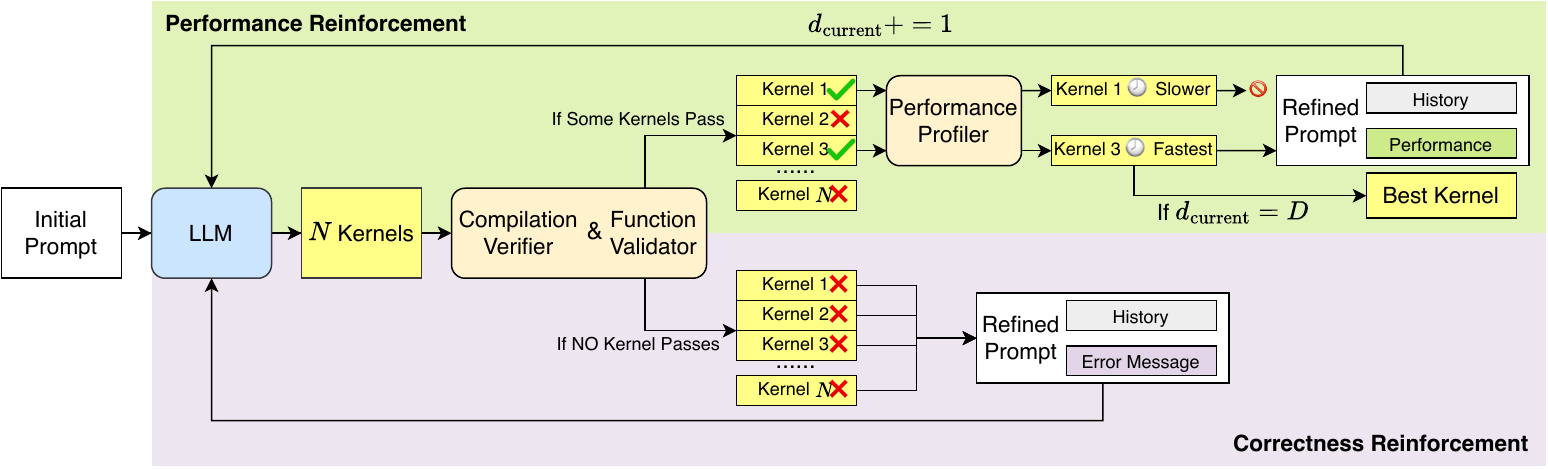}
    \caption{The Search and Reinforcement Process.}
    \label{fig:search_and_reinforcement}
\end{figure*}

\begin{algorithm}[h]
\caption{Feature Search and Reinforcement Framework}
\KwIn{$P_0$ — Initial prompt\; $N$ — Candidates per round\; $D$ — Max depth\;}
\KwOut{$k_{\text{best}}$ — Fastest functionally-correct kernel\;}

$P \leftarrow P_0$\;
$k_{\text{best}} \leftarrow \bot$\; 
$\tau_{\text{best}} \leftarrow \infty$ \tcp*{The best execution time}

$d \leftarrow 1$ \tcp*{The initial depth}
\While{$d \leq D$}{
    \tcp{Generate $N$ kernels}
    $C \leftarrow \text{LLM}(P, N)$\;
    $V \leftarrow \{\,k \in C \mid
        \textsc{CompilationVerifier}(k) \land
        \textsc{FunctionValidator}(k)\,\}$\;

    \uIf{$V = \emptyset$}{
         \tcp{Embed errors \& history}
        $P \leftarrow \textsc{RefinePrompt}(P_0, C)$\;
        \textbf{continue}\;
    }

    $k^\star \leftarrow \mathop{\arg\min}\limits_{k \in V} \left\{\textsc{PerformanceProfiler}(k)\right\}$\;
    $\tau^\star \leftarrow \textsc{PerformanceProfiler}(k^\star)$\;

    \If{$\tau^\star < \tau_{\text{best}}$}{
        $k_{\mathrm{best}} \leftarrow k^\star$\;
        $\tau_{\text{best}} \leftarrow \tau^\star$\;
    }
    \tcp{Add perf hints}
    $d \leftarrow d + 1$\;
    $P \leftarrow \textsc{RefinePrompt}(P_0, k^\star)$\;
}
\Return $k_{\text{best}}$\;
\end{algorithm}

\subsection{Overview}
To facilitate the generation of efficient CUDA kernel code by large language models (LLMs), we propose a novel framework called Feature Search and Reinforcement (FSR). This framework is designed to iteratively explore, evaluate, and optimize CUDA kernels, guiding the LLM toward producing high-performance and functionally correct code tailored to specific hardware environments, as shown in \cref{fig:overview}.
FSR operates based on a multi-prompt paradigm, accepting the following three types of inputs:

\begin{itemize}
\item \textbf{Natural Language:} A high-level \textbf{description} of the target functionality or computational objective that the kernel is expected to achieve.
\item \textbf{Host Code:} A segment of CPU-side \textbf{host code} that defines the context for the CUDA kernel, including declarations of input/output data structures and names.
\item \textbf{GPU Hardware and Architecture Specification:} Detailed information about the target GPU \textbf{hardware} architecture, such as the number of streaming multiprocessors, available shared memory, warp size, and supported compute capabilities.
\end{itemize}

Upon receiving these prompts, FSR orchestrates a series of feature functions to systematically evaluate candidate CUDA kernels. These feature functions assess:

\textbf{Compilation Correctness:} Ensuring the generated kernel code is syntactically valid and successfully compiles.

\textbf{Function Correctness:} Validating the functional correctness of the kernel by comparing its output against expected results or reference implementations.

\textbf{Performance Efficiency:} Measuring execution time and resource utilization to identify execution performance.

Based on this multi-feature evaluation, FSR employs a feedback-driven search strategy to refine and regenerate kernel candidates, progressively reinforcing patterns and features that lead to optimal results. Through this iterative loop of generation, evaluation, and reinforcement, FSR effectively bridges the gap between natural language intent and hardware-optimized CUDA code synthesis.

\subsection{Feature Function}
After the initial CUDA kernel (codes) is generated with the initial prompt, we iteratively evaluate its quality through three evolutionary feature functions: \textbf{Compilation Verifier}, \textbf{Function Validator}, and \textbf{Performance Profiler}. Each function targets a specific aspect of kernel correctness or performance:

\begin{itemize}
    \item \textbf{Compilation Verifier}: This feature function checks whether the generated CUDA kernel can be successfully compiled. Kernels that pass this stage are guaranteed to be free from syntax and compilation errors. Any kernel that fails triggers a detailed compilation error message, which aids in diagnosing and correcting structural issues in the code.

    \item \textbf{Function Validator}: This function evaluates the functional correctness of the kernel. It executes the kernel on a predefined set of test inputs and compares the outputs against expected results. A kernel passes this stage only if its output exactly matches the reference output for all test cases. Any mismatch results in an error message, indicating the presence of semantic or logical flaws.

    \item \textbf{Performance Profiler}: This feature function measures the runtime performance of the kernel on the GPU. It collects execution time and other relevant performance metrics, allowing the framework to assess and compare the efficiency of candidate kernels. Faster kernels are favored in the evolutionary search process.
\end{itemize}

Together, these three feature functions enable the FSR framework to effectively navigate the design space and evolve high-quality CUDA kernels. By enforcing syntactic correctness, validating functional behavior, and profiling runtime performance, the framework ensures that only valid, correct, and efficient kernels are retained for further optimization.

\begin{table*}[]
\centering
\resizebox{0.95\textwidth}{!}{
\begin{tabular}{@{}ccllllllllllllllllllll@{}}
\toprule
GPU & Task ID & 1 & 2 & 3 & 4 & 5 & 6 & 7 & 8 & 9 & 10 & 11 & 12 & 13 & 14 & 15 & 16 & 17 & 18 & 19 & 20 \\ \midrule
\multirow{2}{*}{GTX 1660 SUPER} & LLM (pass@5)  & \textcolor{green}{\Checkmark}  &  \textcolor{green}{\Checkmark} &  \textcolor{green}{\Checkmark} &  \textcolor{green}{\Checkmark} &  \textcolor{green}{\Checkmark} &  \textcolor{red}{\XSolidBrush}  &  \textcolor{green}{\Checkmark} &  \textcolor{green}{\Checkmark} &  \textcolor{green}{\Checkmark} &  \textcolor{green}{\Checkmark}  &  \textcolor{red}{\XSolidBrush}  &   \textcolor{red}{\XSolidBrush}  &   \textcolor{green}{\Checkmark} &  \textcolor{green}{\Checkmark}  &  \textcolor{green}{\Checkmark}  &  \textcolor{red}{\XSolidBrush}  &  \textcolor{green}{\Checkmark}  &  \textcolor{green}{\Checkmark}  &  \textcolor{green}{\Checkmark}  &   \textcolor{green}{\Checkmark} \\
& CUDA-LLM &  \textcolor{green}{\Checkmark} &  \textcolor{green}{\Checkmark} & \textcolor{green}{\Checkmark}  &  \textcolor{green}{\Checkmark} &  \textcolor{green}{\Checkmark} &  \textcolor{green}{\Checkmark} & \textcolor{green}{\Checkmark}  & \textcolor{green}{\Checkmark}  & \textcolor{green}{\Checkmark}  &  \textcolor{green}{\Checkmark}  &  \textcolor{green}{\Checkmark}  &  \textcolor{green}{\Checkmark}  &  \textcolor{green}{\Checkmark}  &  \textcolor{green}{\Checkmark}  &  \textcolor{green}{\Checkmark}  &  \textcolor{green}{\Checkmark}  &  \textcolor{green}{\Checkmark}  &  \textcolor{green}{\Checkmark}  &  \textcolor{green}{\Checkmark}  &   \textcolor{green}{\Checkmark} \\ \midrule

\multirow{2}{*}{RTX 3090 Ti} & LLM (pass@5)  & \textcolor{green}{\Checkmark}  &  \textcolor{green}{\Checkmark} &  \textcolor{green}{\Checkmark} &  \textcolor{green}{\Checkmark} &  \textcolor{green}{\Checkmark} & \textcolor{green}{\Checkmark}  &  \textcolor{green}{\Checkmark} &  \textcolor{green}{\Checkmark} &  \textcolor{green}{\Checkmark} &  \textcolor{green}{\Checkmark}  &  \textcolor{red}{\XSolidBrush}  &  \textcolor{green}{\Checkmark}  &   \textcolor{green}{\Checkmark} &  \textcolor{green}{\Checkmark}  &  \textcolor{green}{\Checkmark}  &  \textcolor{red}{\XSolidBrush}  &  \textcolor{green}{\Checkmark}  &  \textcolor{green}{\Checkmark}  &  \textcolor{green}{\Checkmark}  &   \textcolor{green}{\Checkmark} \\
& CUDA-LLM &  \textcolor{green}{\Checkmark} &  \textcolor{green}{\Checkmark} & \textcolor{green}{\Checkmark}  &  \textcolor{green}{\Checkmark} &  \textcolor{green}{\Checkmark} &  \textcolor{green}{\Checkmark} & \textcolor{green}{\Checkmark}  & \textcolor{green}{\Checkmark}  & \textcolor{green}{\Checkmark}  &  \textcolor{green}{\Checkmark}  &  \textcolor{green}{\Checkmark}  &  \textcolor{green}{\Checkmark}  &  \textcolor{green}{\Checkmark}  &  \textcolor{green}{\Checkmark}  &  \textcolor{green}{\Checkmark}  &  \textcolor{green}{\Checkmark}  &  \textcolor{green}{\Checkmark}  &  \textcolor{green}{\Checkmark}  &  \textcolor{green}{\Checkmark}  &   \textcolor{green}{\Checkmark} \\ \bottomrule
\end{tabular}
}
\caption{Comparison of functional correctness between directly generated CUDA kernels and FSR-optimized kernels. Each Task ID corresponds to the task described in Table~\ref{tab:task}.}
\label{tab:correctness}
\end{table*}

\subsection{Search and Reinforcement}

Based on the evaluation and feedback provided by our designed feature functions, we optimize the CUDA kernels generated by the LLM through iterative refinement of the input prompt. The optimization process consists of two stages: ensuring correctness and improving performance, as shown in Fig.~\ref{fig:search_and_reinforcement}. Initially, the LLM generates $N$ candidate CUDA kernels from an initial prompt. Each candidate is sequentially evaluated by two correctness checking modules: the \textbf{Compilation Verifier} and the \textbf{Function Validator}. Suppose none of the candidates pass these checks. In that case, the FSR framework constructs a new prompt by incorporating the current kernel code, the corresponding error messages, and the interaction history with the LLM. It uses this refined prompt to generate a new set of $N$ candidate kernels. This process repeats until at least one candidate successfully compiles and produces correct output. These valid kernels are then passed to the \textbf{Performance Profiler}, which measures their execution speed on the GPU. The fastest kernel is selected, and its code, along with the associated performance-optimized prompt and dialogue history, forms a new prompt for the next round of LLM generation. This cycle of correctness verification and performance optimization continues until a predefined depth $D$ is reached. The kernel with the highest execution efficiency at the final iteration is returned as the final output of the FSR framework.

FSR effectively assists LLMs in overcoming the challenges of CUDA kernel generation by continuously evaluating various kernel features, thereby progressively reinforcing LLMs’ capability to produce efficient CUDA kernels. Through the proposed multi-dimensional validation process, FSR successfully addresses two critical objectives of ensuring correctness and optimizing runtime performance.

\begin{table*}
    \centering
    \begin{tabularx}{0.9\textwidth}{ccX}
    \toprule
    \textbf{Task ID} & \textbf{Task} & \multicolumn{1}{c}{\textbf{Description}}  \\
       \midrule
1 & Sigmoid & Applies the sigmoid activation function to each element in the input array: $\text{sigmoid}(x) = 1 / (1 + \exp(-x))$. \\ \midrule
2 & Matrix multiplication & Performs parallel multiplication of two matrices, accelerating large-scale matrix operations.  \\ \midrule
3 & Max Pooling 3D & Slides a 3D window and keeps only the maximum value in each block, down-sampling the volume. \\ \midrule
4 & LayerNorm & Normalizes activations per sample, then applies learned scale $\gamma$ and bias $\beta$. \\ \midrule
5 & 2D Convolution & Applies a 2D convolution filter over the input matrix (e.g., image) to extract spatial features. \\ \midrule
6 & Multi-Head Self-Attention & Splits the input into multiple parts, computes attention in parallel for each part, then combines the results. This helps the model focus on different types of information at once. \\ \midrule
7 & Mean Square Error & Computes the mean squared error between predicted and target values for regression tasks. \\ \midrule
8 & Matrix Transpose & Transposes the input matrix by swapping rows and columns in parallel. \\ \midrule
9 & Reverse Array & Reverses the order of elements in a one-dimensional array using parallel operations. \\ \midrule
10 & ReLU Activation Fuction & Applies the ReLU function element-wise: $\text{ReLU}(x)=\max(0,x)$, commonly used in neural networks. \\ \midrule
11 & Top-K Selection & Selects the top $k$ largest or smallest elements from an input array in parallel. \\ \midrule
12 & Sorting & Sorts an array in ascending or descending order using a parallel sorting algorithm.  \\ \midrule
13 & Matrix Copy & Copies matrix data from a source to a destination in parallel, enabling fast memory transfer.  \\ \midrule
14 & Reduction & Aggregates elements of an array (e.g., sum, max, min) into a single value using parallel tree-based reduction. \\ \midrule
15 & Dot Product & Computes the dot product of two vectors by multiplying corresponding elements and summing the results in parallel. \\ \midrule
16 & Prefix Sum & Produces the cumulative sum of an array in parallel. \\ \midrule
17 & Categorical Cross-Entropy Loss & Calculates the cross-entropy loss between predicted probability distributions and one-hot encoded labels, often used in classification tasks. \\ \midrule
18 & Monte Carlo Integration & Estimates the integral of a function using random sampling and averaging, leveraging GPU parallelism for high sampling efficiency. \\ \midrule
19 & Histogramming & Tallies how many inputs fall into each bin to form a histogram.  \\ \midrule
20 & Ordinary Least Squares Regression & Solves closed-form solution $(X^TX)^{-1}X^Ty$ to the Ordinary Least Squares (OLS) regression problem for choosing the unknown parameters in a linear regression model. \\ \bottomrule
\end{tabularx}
\caption{Description and features of evaluation benchmark tasks.}
\label{tab:task}
\end{table*}

\begin{figure*}[t]
    \centering
    \begin{subfigure}[t]{0.47\linewidth}
        \centering
        \includegraphics[width=\linewidth]{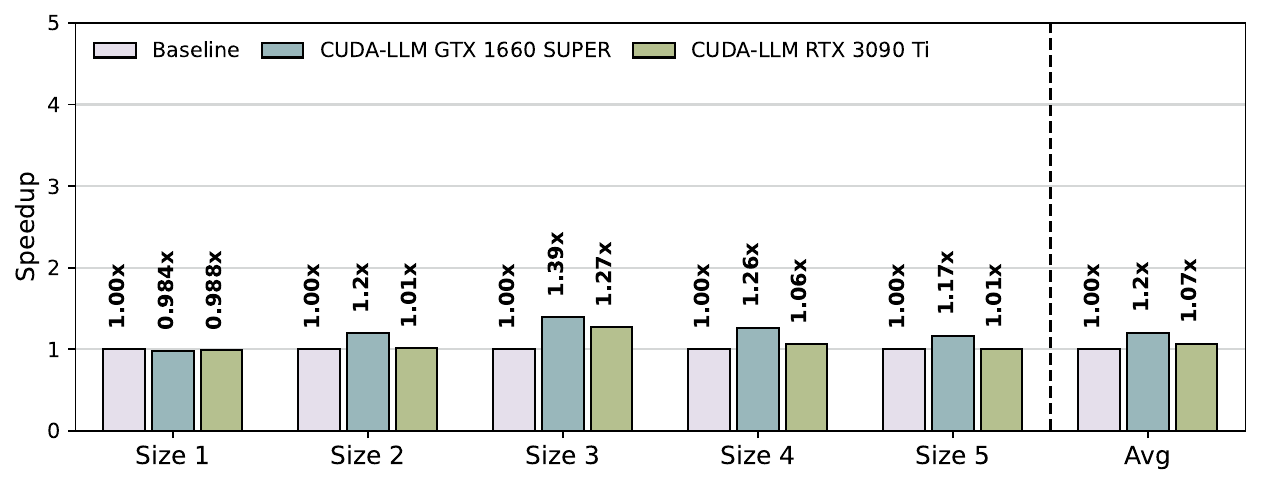}
        \caption{Task 1: Sigmoid}
        \label{fig:task1}
    \end{subfigure}%
    \begin{subfigure}[t]{0.47\linewidth}
        \centering
        \includegraphics[width=\linewidth]{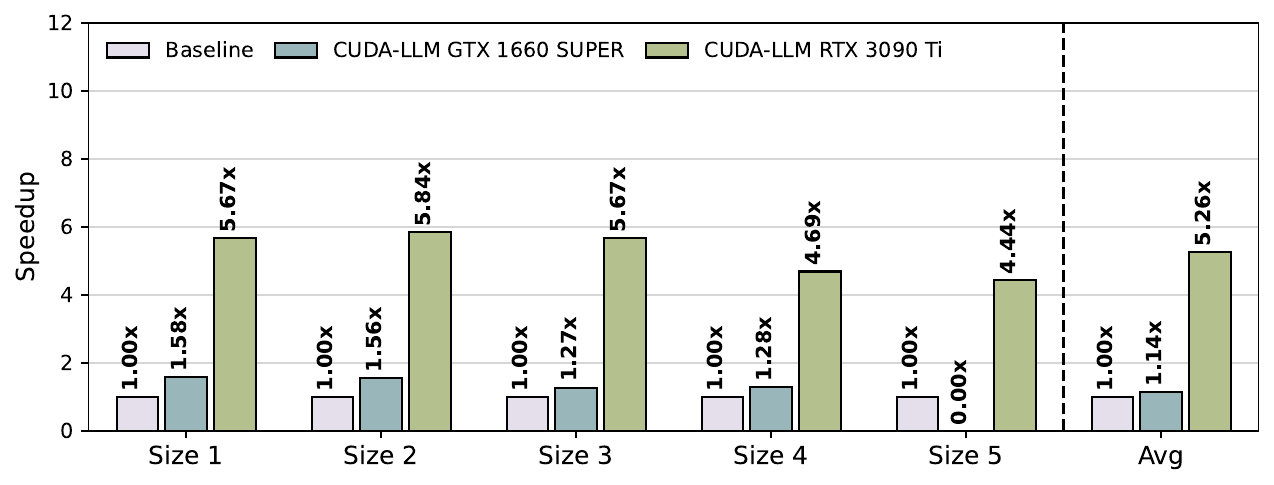}
        \caption{Task 2: Matrix Multiplication}
        \label{fig:task2}
    \end{subfigure}

    \begin{subfigure}[t]{0.47\linewidth}
        \centering
        \includegraphics[width=\linewidth]{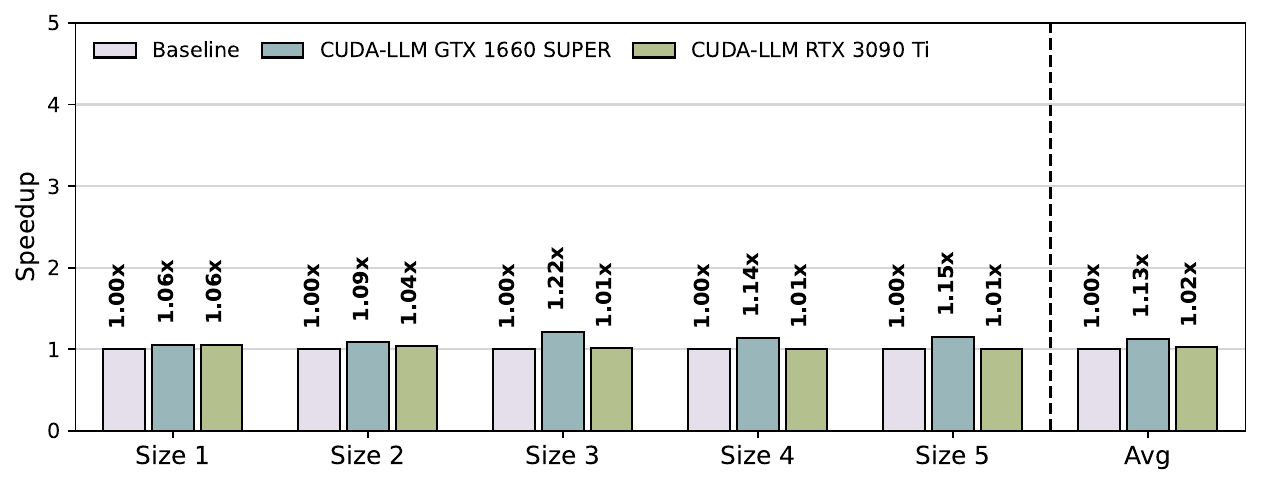}
        \caption{Task 3: Max Pooling 3D}
        \label{fig:task3}
    \end{subfigure}%
    \begin{subfigure}[t]{0.47\linewidth}
        \centering
        \includegraphics[width=\linewidth]{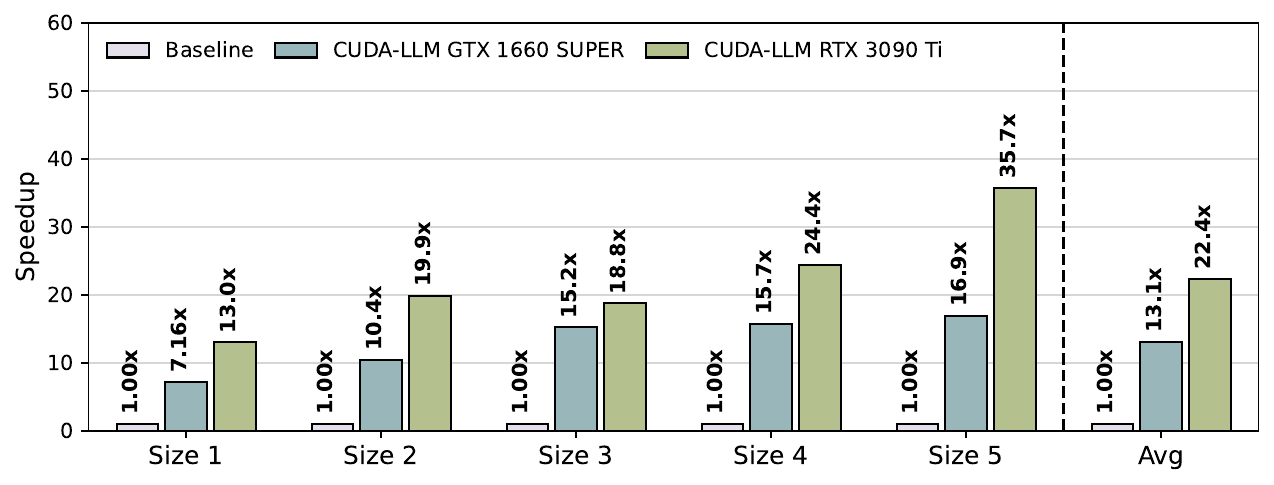}
        \caption{Task 4: LayerNorm}
        \label{fig:task4}
    \end{subfigure}

    \begin{subfigure}[t]{0.47\linewidth}
        \centering
        \includegraphics[width=\linewidth]{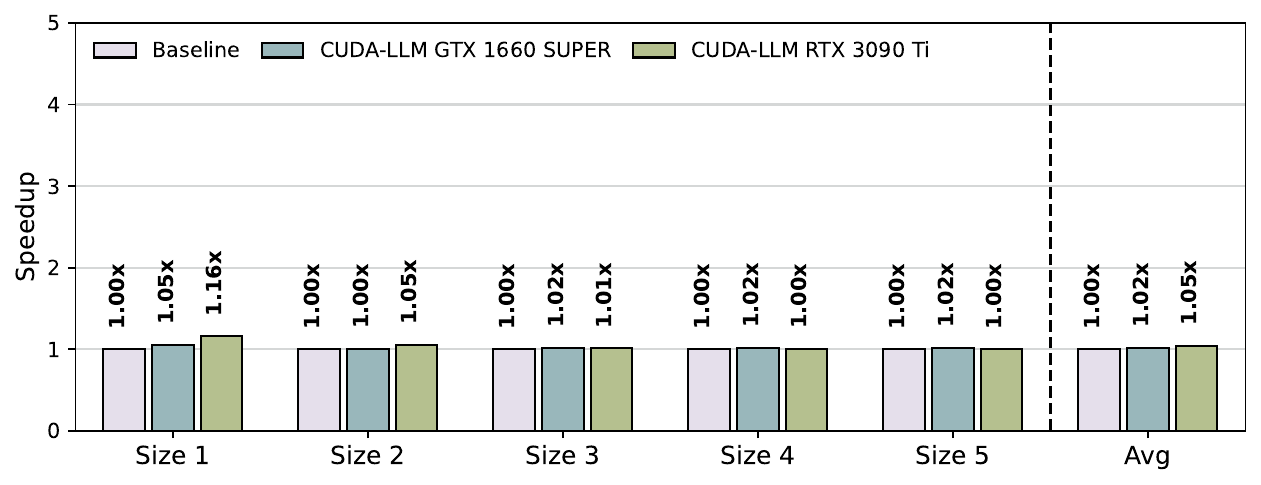}
        \caption{Task 5: 2D Convolution}
        \label{fig:task5}
    \end{subfigure}%
    \begin{subfigure}[t]{0.47\linewidth}
        \centering
        \includegraphics[width=\linewidth]{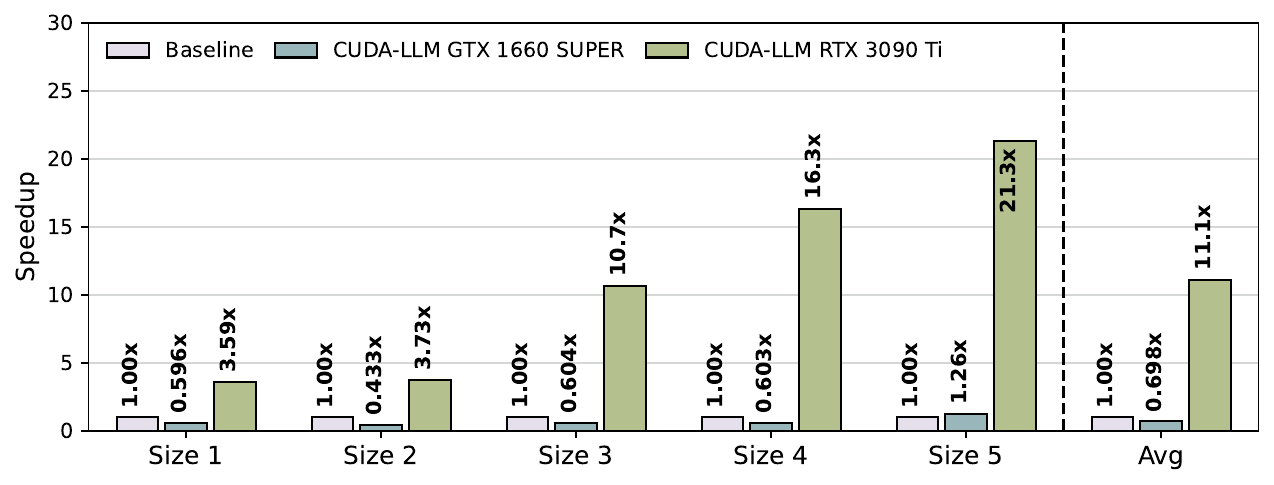}
        \caption{Task 6: Multi-Head Self-Attention}
        \label{fig:task6}
    \end{subfigure}

    \begin{subfigure}[t]{0.47\linewidth}
        \centering
        \includegraphics[width=\linewidth]{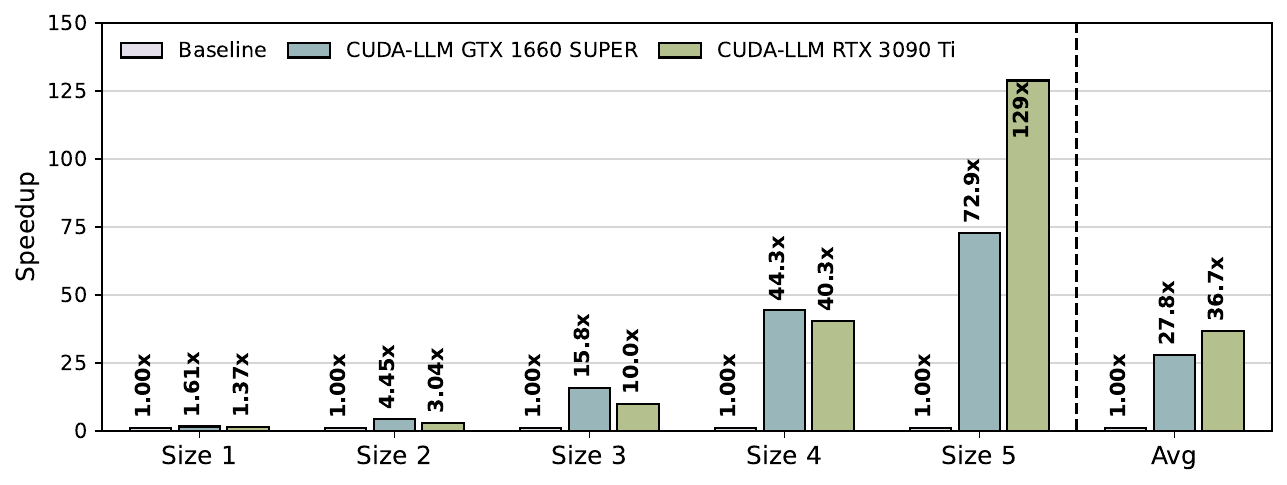}
        \caption{Task 7: Mean Square Error}
        \label{fig:task7}
    \end{subfigure}%
    \begin{subfigure}[t]{0.47\linewidth}
        \centering
        \includegraphics[width=\linewidth]{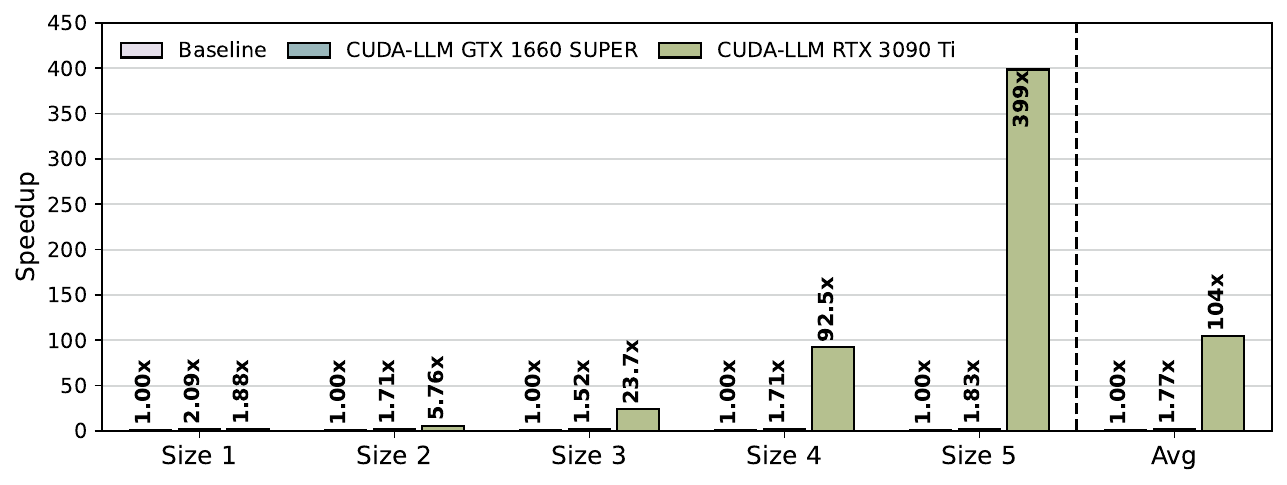}
        \caption{Task 8: Matrix Transpose}
        \label{fig:task8}
    \end{subfigure}

    \begin{subfigure}[t]{0.47\linewidth}
        \centering
        \includegraphics[width=\linewidth]{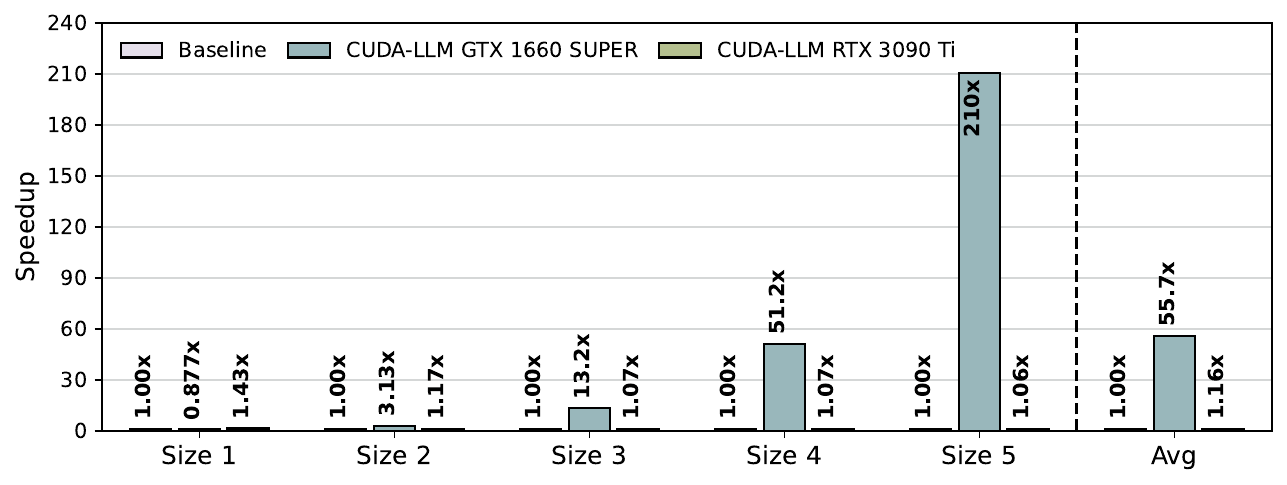}
        \caption{Task 9: Reverse Array}
        \label{fig:task9}
    \end{subfigure}%
    \begin{subfigure}[t]{0.47\linewidth}
        \centering
        \includegraphics[width=\linewidth]{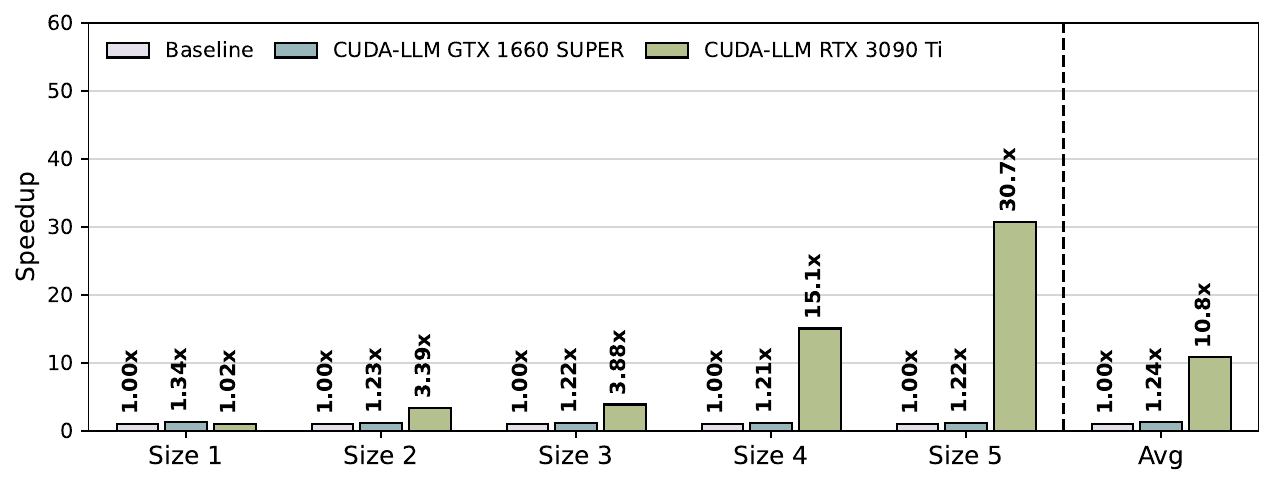}
        \caption{Task 10: ReLU Activation Fuction}
        \label{fig:task10}
    \end{subfigure}

    \begin{subfigure}[t]{0.47\linewidth}
        \centering
        \includegraphics[width=\linewidth]{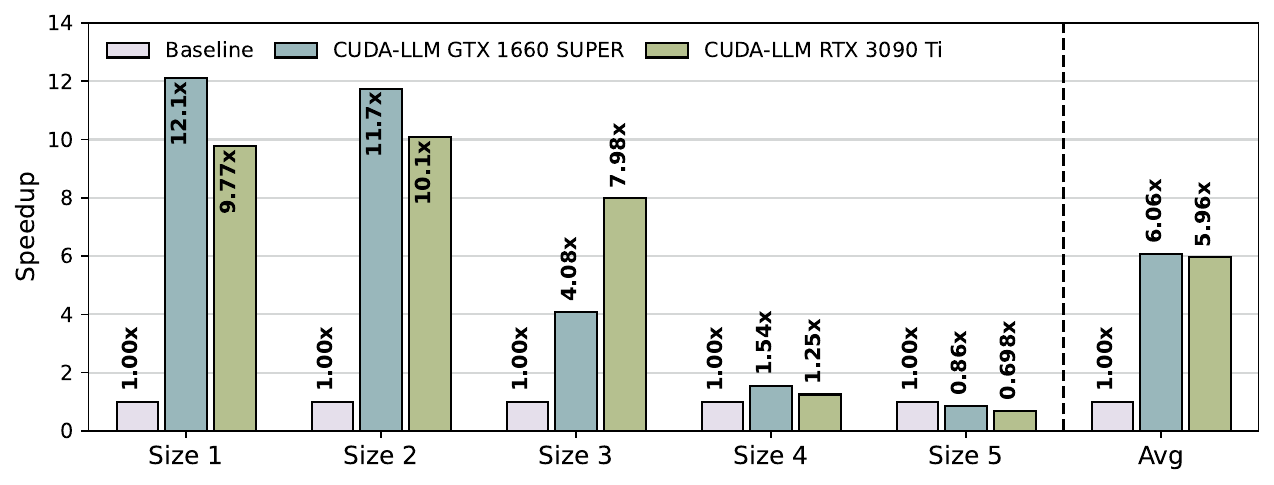}
        \caption{Task 11: Top-K Selection}
        \label{fig:task11}
    \end{subfigure}%
    \begin{subfigure}[t]{0.47\linewidth}
        \centering
        \includegraphics[width=\linewidth]{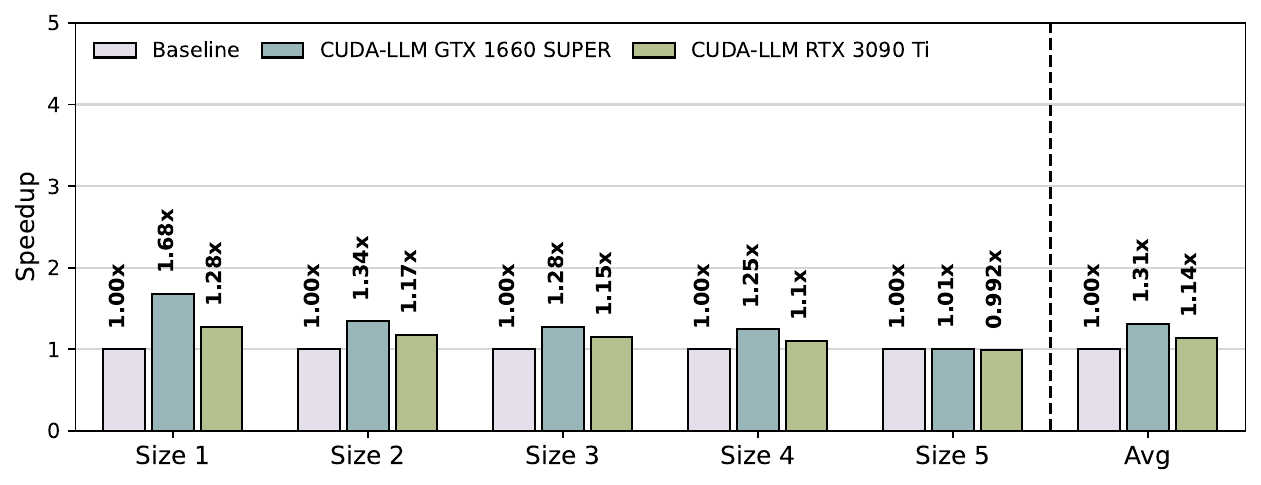}
        \caption{Task 12: Sorting}
        \label{fig:task12}
    \end{subfigure}

\end{figure*}

\begin{figure*}[t] \ContinuedFloat
    \centering

    \begin{subfigure}[t]{0.47\linewidth}
        \centering
        \includegraphics[width=\linewidth]{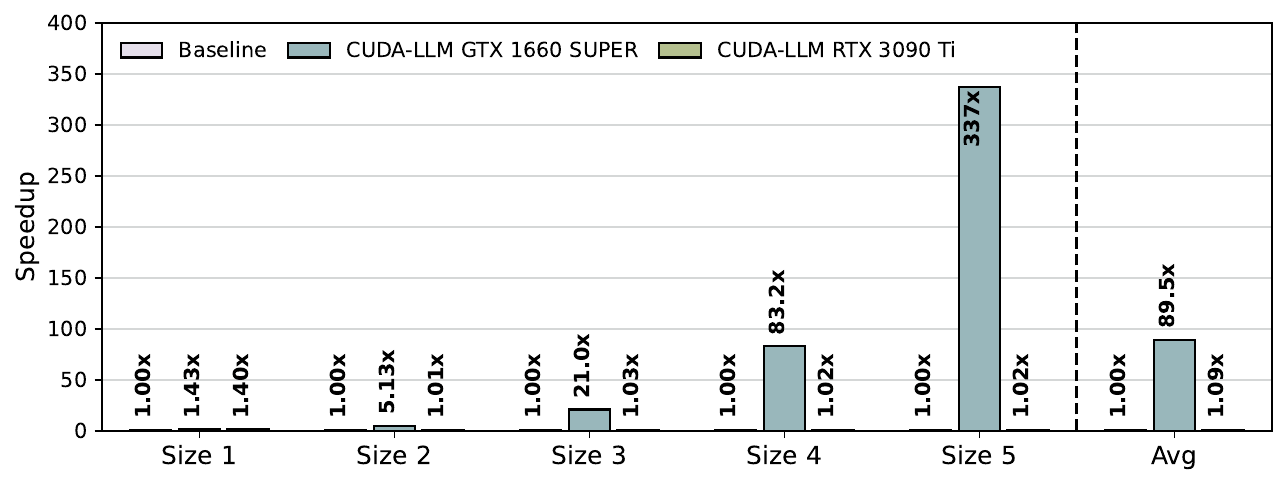}
        \caption{Task 13: Matrix Copy}
        \label{fig:task13}
    \end{subfigure}%
    \begin{subfigure}[t]{0.47\linewidth}
        \centering
        \includegraphics[width=\linewidth]{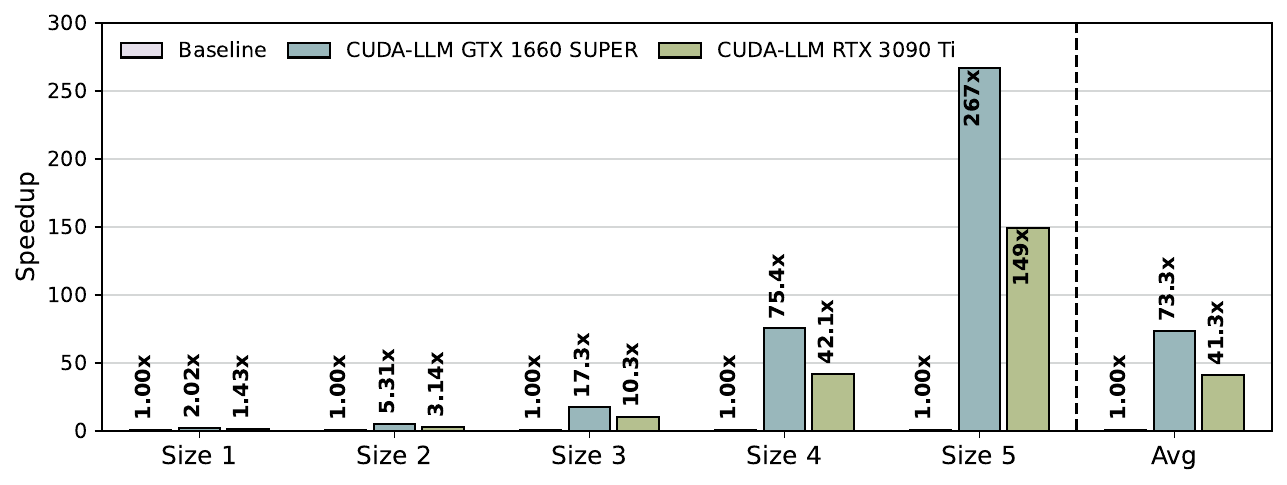}
        \caption{Task 14: Reduction}
        \label{fig:task14}
    \end{subfigure}
    
    \begin{subfigure}[t]{0.47\linewidth}
        \centering
        \includegraphics[width=\linewidth]{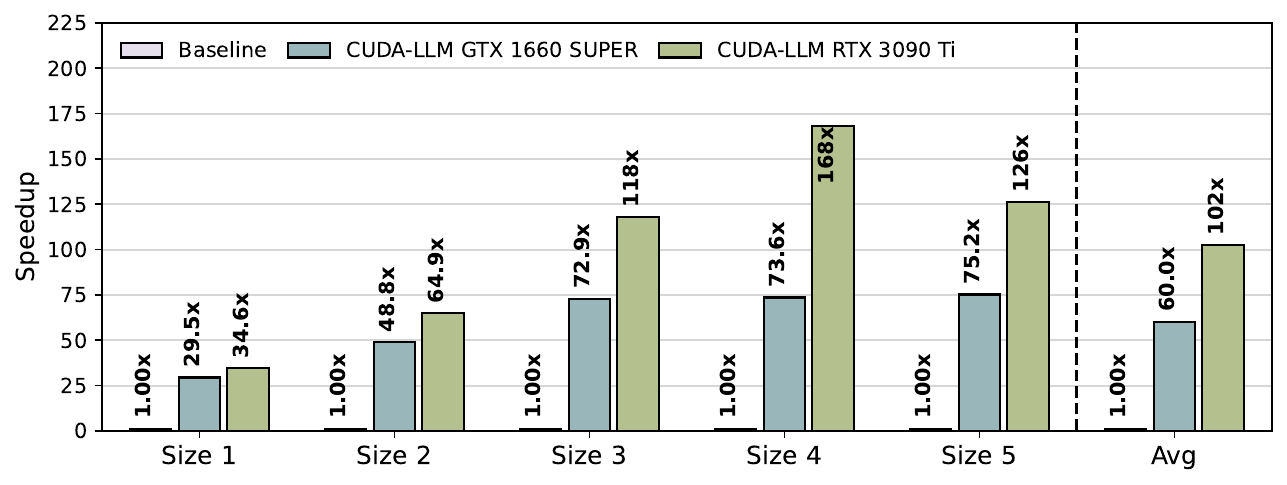}
        \caption{Task 15: Dot Product}
        \label{fig:task15}
    \end{subfigure}%
    \begin{subfigure}[t]{0.47\linewidth}
        \centering
        \includegraphics[width=\linewidth]{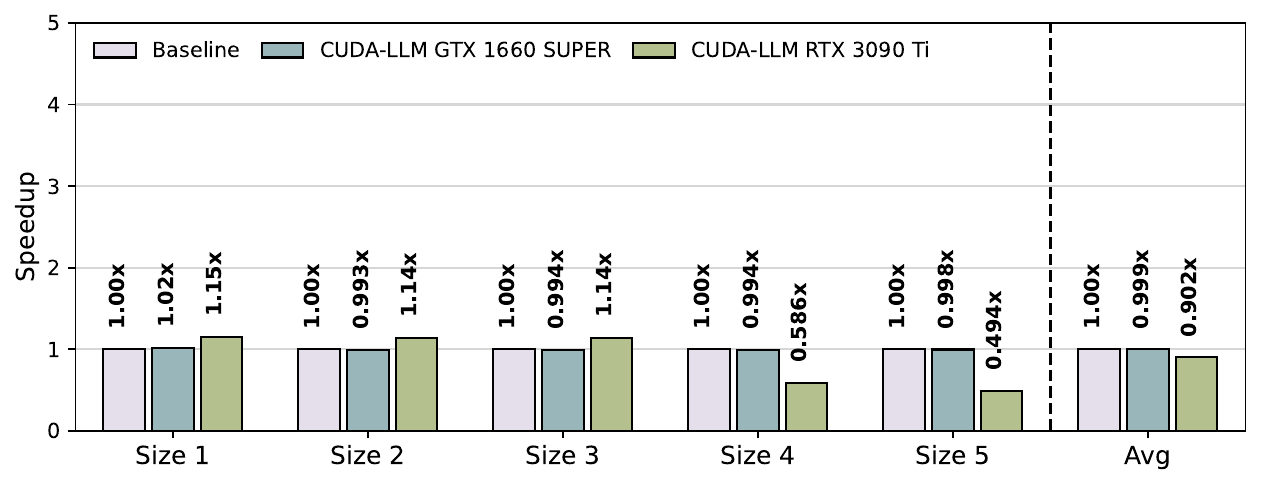}
        \caption{Task 16: Prefix Sum}
        \label{fig:task16}
    \end{subfigure}

    \begin{subfigure}[t]{0.47\linewidth}
        \centering
        \includegraphics[width=\linewidth]{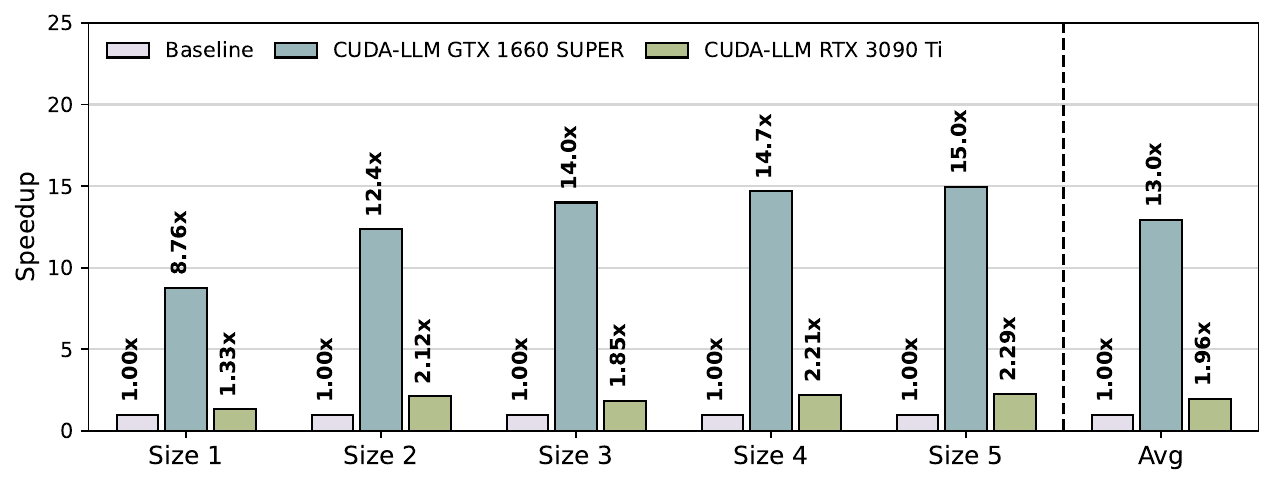}
        \caption{Task 17: Categorical Cross-Entropy Loss}
        \label{fig:task17}
    \end{subfigure}%
    \begin{subfigure}[t]{0.47\linewidth}
        \centering
        \includegraphics[width=\linewidth]{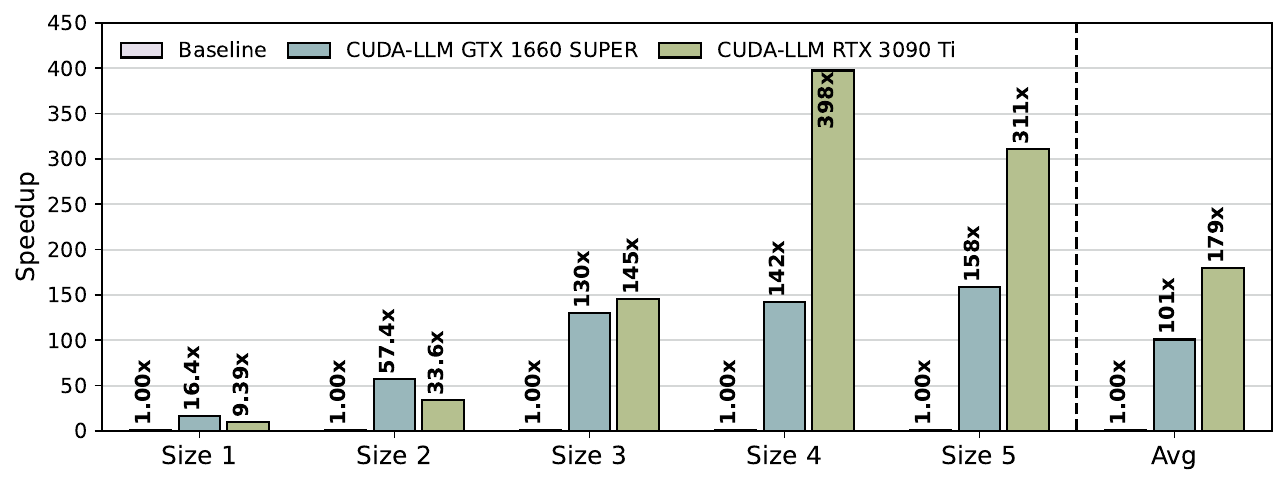}
        \caption{Task 18: Monte Carlo Integration}
        \label{fig:task18}
    \end{subfigure}

    \begin{subfigure}[t]{0.47\linewidth}
        \centering
        \includegraphics[width=\linewidth]{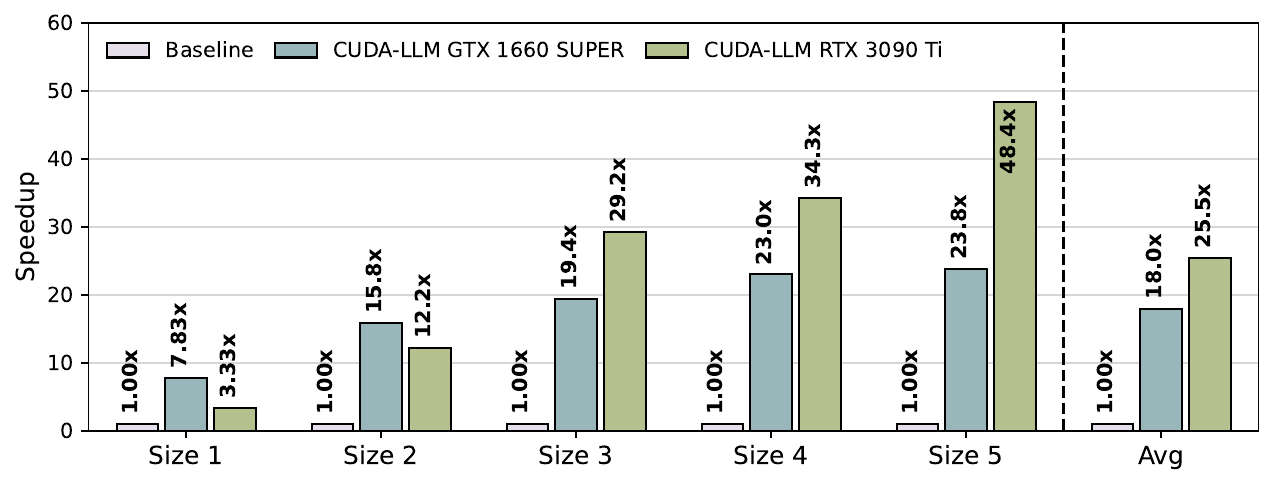}
        \caption{Task 19: Histogramming}
        \label{fig:task19}
    \end{subfigure}%
    \begin{subfigure}[t]{0.47\linewidth}
        \centering
        \includegraphics[width=\linewidth]{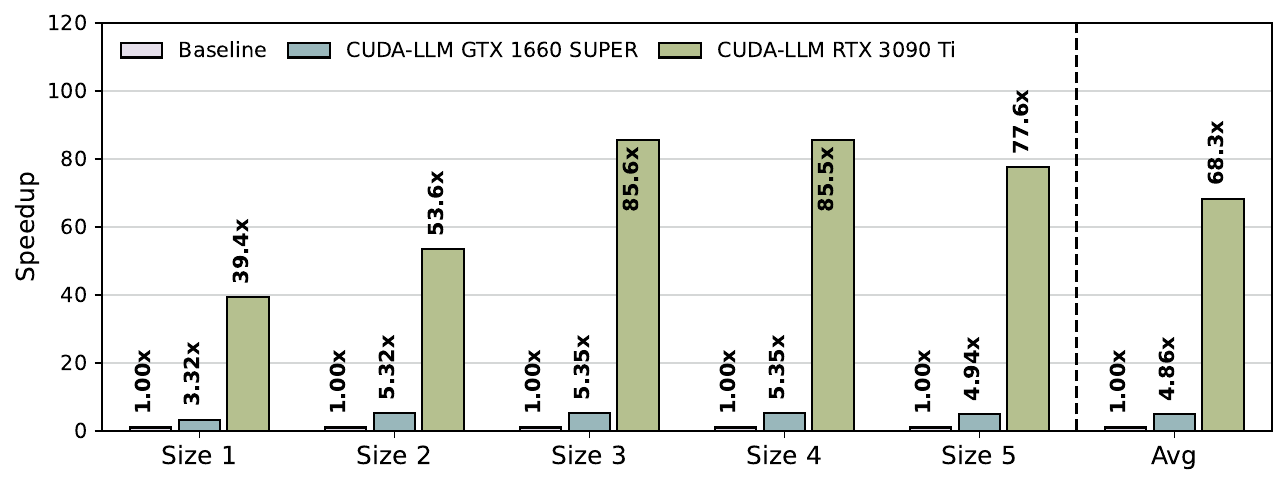}
        \caption{Task 20: Ordinary Least Squares Regression}
        \label{fig:task20}
    \end{subfigure}

    \caption{Performance evaluation across various tasks.}
    \label{fig:speedup}
\end{figure*}

\section{Evaluation}
\label{sec:evaluation}
We present the results of CUDA-LLM in this section. Our experiments demonstrate that FSR can protect the correctness and bring a significant performance gain across various tasks.
\subsection{Evaluation Setup}

\subsubsection{Benchmarks and Evaluation Metrics}  
We evaluate our proposed method using a diverse set of benchmarks comprising 20 widely-used GPU kernel functions, selected from three sources: the official NVIDIA CUDA Samples \cite{nvidia2025cudasamples}, LeetGPU \cite{leetgpu_challenges}, and KernelBench \cite{ouyang2025kernelbench}. These benchmarks span a variety of computational patterns and memory access behaviors, providing a representative evaluation of real-world CUDA kernel workloads. The specific tasks and their characteristics are summarized in Table~\ref{tab:task}. For each kernel, we test across variable input data sizes (detailed in Table~\ref{tab:input_size}) to assess the scalability and robustness of our method under different workload intensities. 

We evaluate performance using the following metrics:
\begin{itemize}
    \item \textbf{Correctness}: Incoperating both compilation and function correctness. We consider the result correct if all elements of the output differ from the human-written CUDA results by no more than a pre-determined deviation.
    \item \textbf{Runtime Latency}: The measured execution time of each kernel on GPUs and the speedup is normalized to the human-written CUDA baseline.
\end{itemize}

\subsubsection{Backbone Model}  
We employ \textbf{DeepSeek-V3-0324}~\cite{liu2024deepseek}, a state-of-the-art large language model, as the core inference engine in our methodology. This model is chosen for its demonstrated capability in code understanding, reasoning, and performance prediction, making it well-suited for tasks involving CUDA kernel analysis and optimization recommendations.

\subsubsection{Hardware Setup}  
The evaluations are conducted on two representative GPU platforms to analyze the performance of our method across different hardware configurations:
\begin{itemize}
    \item \textbf{Edge Platform}: NVIDIA GeForce GTX 1660 SUPER GPU with the Turing architecture. This setup reflects a resource-constrained edge computing environment.
    \item \textbf{Server Platform}: NVIDIA GeForce RTX 3090 Ti GPU with the Ada Lovelace architecture. This high-end configuration represents a server-grade computing environment with ample computational and memory resources.
\end{itemize}

This dual-platform setting allows us to investigate how well the CUDA-LLM generalizes and scales under both edge and high-performance GPU scenarios.


\subsection{Evaluation Results}
\subsubsection{Compilation and Function Correctness}

In the experiments, the LLM is instructed to generate $N$ CUDA kernels directly, and a task is considered successful if at least one candidate passes the Function Validator. As shown in Table~\ref{tab:correctness}, under the CUDA-LLM FSR framework, the LLM successfully completed all predefined tasks on GTX 1660 SUPER GPU and RTX 3090 Ti GPU. In contrast, when generating code directly without FSR, the LLM produces incorrect outputs, with all $N$ candidates failing to meet the correctness criteria in some tasks. 

The FSR in CULDA-LLM not only optimizes the initial kernel but also addresses new accuracy issues that may arise in LLM-generated kernels after performance optimization. FSR continuously refines the code to resolve these issues. For example, in Task 11, shared-memory is used to optimize 2D convolution; however, the kernel silently produced incorrect results due to missing corner halo data, leading to inaccurate outputs despite successful execution. FSR automatically detected the mismatch with reference outputs and corrected the kernel by replacing the faulty shared-memory logic with a fully correct global memory-based implementation. This ensured consistent numerical accuracy across all input sizes, demonstrating FSR’s effectiveness in not only optimizing performance but also guaranteeing functional correctness.

\subsubsection{Execution Performance}
\cref{fig:speedup} demonstrates the execution speedup achieved by the CUDA kernels generated using our proposed FSR framework across twenty commonly used tasks, evaluated under five test input sizes (from Size 1 to Size 5, in increasing order detailed in Table~\ref{tab:input_size}). The results show that, compared to directly generated CUDA kernels, FSR consistently improves execution efficiency across nearly all tasks. In certain tasks, FSR achieves remarkable performance gains. As shown in \cref{fig:task8} (Task 8: Matrix Transpose), \cref{fig:task15} (Task 15: Dot Product), and \cref{fig:task18} (Task 18: Monte Carlo Integration), the CUDA kernels generated by FSR achieve average speedups of 104$\times$, 102$\times$, and 179$\times$ respectively, compared to the baseline on an RTX 3090 Ti GPU. Similarly, on Tasks 13 and 14 (\cref{fig:task13,fig:task14}), FSR delivers 89.5$\times$ and 73.3$\times$ speedups over the baseline on a GTX 1660 SUPER GPU. Notably, for some large input sizes in the Matrix Transpose, Monte Carlo Integration, and Matrix Copy tasks, the speedup exceeds 300$\times$. While the speedups in most tasks are less dramatic than those mentioned above, FSR still delivers substantial performance improvements, achieving effective accelerations ranging from 5.95$\times$ to 68.3$\times$ on RTX 3090 Ti GPU and 6.06$\times$ to 60.0$\times$ on GTX 1660 SUPER GPU (\cref{fig:task2,fig:task4,fig:task7,fig:task10,fig:task11,fig:task14,fig:task19,fig:task20}). Naturally, our experiments also revealed that in a few tasks, the performance improvement achieved by FSR was less pronounced.

By analyzing the kernels generated under the FSR framework and comparing them to the baseline kernels, we observed that the optimized versions incorporate a wide range of effective techniques. In the Matrix Transpose task, for instance, the baseline kernel performs one scattered global read and one scattered global write per element; its write pattern (column-major) defeats coalescing, so memory bandwidth drops to a few GB/s. The FSR-generated version batches a large number ($\verb|TILE_DIM*TILE_DIM|$) of elements per 32-thread warp, turns both accesses into fully coalesced, conflict-free bursts, and cuts the number of global transactions nearly in half. Combined with loop unrolling (\verb|#pragma unroll|) and lighter address arithmetic, DRAM bandwidth is saturated and arithmetic latency is hidden, yielding the observed 104$\times$ speedup on modern GPUs. Moreover, the performance gains become more pronounced as the input data size increases (\cref{fig:task8}). Additionally, many tasks — such as Reduction and Monte Carlo Integration — benefit significantly from the use of warp-level primitives. Warp-level primitives enhance efficiency by enabling direct communication and data sharing among threads within the same warp, reducing the need for slower global or shared memory access. This leads to lower synchronization overhead and improved parallelism, particularly in operations such as Reduction and Monte Carlo Integration where intermediate results must be aggregated or exchanged efficiently.


\section{Conclusion}
\label{sec:conclusion}
Through the proposed Feature Search and Reinforcement (FSR) framework, we demonstrate that by integrating feature-aware search with reinforcement-driven refinement, FSR enhances the capabilities of LLMs from producing generic code to generating intelligent, hardware-specific, architecture-aware, and performance-critical implementations. This advancement empowers LLMs to effectively navigate the complexities of parallel computing architectures and platform-specific constraints. Empirical evaluations highlight the effectiveness of FSR in CUDA-LLM, showcasing its potential to enable the next generation of automated, high-performance code generation tools for GPUs and other heterogeneous computing platforms.

\bibliographystyle{ieeenat_fullname}
\balance
\bibliography{bio}
\onecolumn
\setcounter{page}{1}
\appendix

\section*{Appendix}

\section{Size Settings of Test Inputs}
\label{sec:size}
Table \ref{tab:input_size} provides a detailed overview of the input data sizes used to evaluate the functional correctness of the kernel in the Function Validator.

\begin{table*}[!ht]
    \centering
    \begin{tabularx}{\textwidth}{ccX}
    \toprule
    \multicolumn{1}{c}{\textbf{Task ID}} & \multicolumn{1}{c}{\textbf{Task}} & \multicolumn{1}{c}{\textbf{Input Size}}  \\
       \midrule
1 & Sigmoid & $(16, \text{dim})$, $\text{dim}=\{2^{10}, 2^{12}, 2^{14}, 2^{16}, 2^{18}\}$ \\ \midrule
\multirow{2}{*}{2} & \multirow{2}{*}{Matrix Multiplication} & $A$: $(\text{size}, 4096)$, $\text{size}=\{2^{10}, 2^{12}, 2^{14}, 2^{16}, 2^{18}\}$ \\
                  &                   &  $B$: $(4096, 2048)$ \\ \midrule
3 & Max Pooling 3D & $(16, 32, \text{dim1}, \text{dim2}, \text{dim3})$, $\text{dim1}=\text{dim2}=\text{dim3}=\{16, 24, 32, 40, 48\}$ \\ \midrule
4 & LayerNorm & $(16, 4, \text{dim1}, \text{dim2})$, $\text{dim1}=\text{dim2}=\{2^{6}, 2^{7}, 2^{8}, 2^{9}, 2^{10}\}$ \\ \midrule
\multirow{2}{*}{5} & \multirow{2}{*}{2D Convolution} & 2D matrix: $(\text{size}, \text{size})$, $\text{size}=\{128, 256, 512, 1024, 2048\}$ \\ 
  &     &  kernel: $(24, 24)$ \\ \midrule
6 & Multi-Head Self-Attention & $\{(N, d_\text{model}, h)\}=\{(64, 32, 4), (128,  64, 8), (256, 128, 8), (384, 256, 16),$\\& & $(512, 512, 16)\}$ \\ \midrule
\multirow{2}{*}{7} & \multirow{2}{*}{Mean Square Error} & predictions: $N$, $N=\{2^{10}, 2^{12}, 2^{14}, 2^{16}, 2^{18}\}$ \\ 
&  & targets: $N$ \\ \midrule
8 & Matrix Transpose  & $A$: $(N, N)$, $N=\{2^{10}, 2^{11}, 2^{12}, 2^{13}, 2^{14}\}$ \\ \midrule
9 & Reverse Array & $N$, $N=\{2^{20}, 2^{22}, 2^{24}, 2^{26}, 2^{28}\}$ \\ \midrule
10 & ReLU Activation Fuction &  $N$, $N=\{2^{20}, 2^{22}, 2^{24}, 2^{26}, 2^{28}\}$ \\ \midrule
11 & Top-K Selection & $\{(N, k)\}=\{(2^{10}, 32), (2^{12}, 64), (2^{14}, 128), (2^{16}, 256), (2^{18}, 512)\}$ \\ \midrule
12 & Sorting & $N$, $N=\{2^{9}, 2^{10}, 2^{11}, 2^{12}, 2^{13}\}$ \\ \midrule
13 & Matrix Copy & $A$: $(N, N)$, $N=\{2^{10}, 2^{11}, 2^{12}, 2^{13}, 2^{14}\}$ \\ \midrule
14 & Reduction & $N$, $N=\{2^{10}, 2^{12}, 2^{14}, 2^{16}, 2^{18}\}$ \\ \midrule
\multirow{2}{*}{15} & \multirow{2}{*}{Dot Product} & $A$: $N$, $N=\{2^{16}, 2^{17}, 2^{18}, 2^{19}, 2^{20}\}$ \\
                  &                   &  $B$: $N$ \\ \midrule
16 & Prefix Sum & $N$, $N=\{2^{10}, 2^{12}, 2^{14}, 2^{16}, 2^{18}\}$ \\ \midrule
\multirow{2}{*}{17} & \multirow{2}{*}{Categorical Cross-Entropy Loss} & $N=\{2^{14}, 2^{16}, 2^{18}, 2^{20}, 2^{22}\}$ \\ 
& & $C=10$ \\ \midrule
\multirow{3}{*}{18} & \multirow{3}{*}{Monte Carlo Integration} & $a=0$, $b=1$ \\
&  & $f(x)=\sin{(2\pi x)}$ \\
&  & $N=\{2^{14}, 2^{16}, 2^{18}, 2^{20}, 2^{22}\}$ \\ \midrule
\multirow{2}{*}{19} & \multirow{2}{*}{Histogramming} & $\text{num\_bins}=256$ \\ 
& & $N=\{2^{16}, 2^{18}, 2^{20}, 2^{22}, 2^{24}\}$ \\ \midrule
\multirow{2}{*}{20} & \multirow{2}{*}{Ordinary Least Squares Regression} & $n\_samples=\{2^{14}, 2^{16}, 2^{18}, 2^{20}, 2^{22}\}$ \\
& & $n\_features=10$ \\ \bottomrule
\end{tabularx}
\caption{Size settings of randomly generated test inputs used in the Function Validator.}
\label{tab:input_size}
\end{table*}

\section{Prompt}
\label{sec:prompt_text}

\subsection{Initial Prompt}
\begin{tcolorbox}[title=\textbf{Prompt},colback=SeaGreen!10!CornflowerBlue!10,colframe=RoyalPurple!55!Aquamarine!100!]
Write a CUDA kernel function on [Device type] GPU, utilizing the functions described as below:

[Task]: [Prompt]

The output should be the content of whole .cu file containing ONE kernel function, completing the reference code below:

[Code]

Do not modify the test part.
\end{tcolorbox}

\subsection{Refined Prompt}
When the generated $N$ candidates contain at least one valid kernel that passes both the Compilation Verifier and the Function Validator, the Performance Profiler selects the fastest kernel and constructs a new Refined Prompt for the subsequent round of LLM generation. The format of this Refined Prompt is as follows:

\begin{tcolorbox}[title=\textbf{Prompt},colback=SeaGreen!10!CornflowerBlue!10,colframe=RoyalPurple!55!Aquamarine!100!]
Optimize the kernel function for less execution time on [Device type] GPU.

The output should be the content of whole .cu file containing ONE kernel function.

Do not modify the test part.
\end{tcolorbox}

When none of the generated $N$ candidates pass both the Compilation Verifier and the Function Validator, the FSR framework constructs a new prompt for each candidate. If a candidate fails specifically at the Compilation Verifier stage, the corresponding Refined Prompt is structured as follows:

\begin{tcolorbox}[title=\textbf{Prompt},colback=SeaGreen!10!CornflowerBlue!10,colframe=RoyalPurple!55!Aquamarine!100!]
Modify the code with the execution error result.

The output should be the content of whole .cu file containing ONE kernel function.

Do not modify the test part.

The execution output is:

[Execution error output]
\end{tcolorbox}

When none of the generated $N$ candidates pass both the Compilation Verifier and the Function Validator, the FSR framework constructs a new prompt for each candidate. If a candidate fails at the Function Validator specifically due to incorrect kernel execution (despite successful compilation), the corresponding Refined Prompt is structured as follows:

\begin{tcolorbox}[title=\textbf{Prompt},colback=SeaGreen!10!CornflowerBlue!10,colframe=RoyalPurple!55!Aquamarine!100!]
The code failed to launch the kernel. Modify the code with the device information: [Device type] GPU.

The output should be the content of whole .cu file containing ONE kernel function.

Do not modify the test part.
\end{tcolorbox}

When none of the generated $N$ candidates pass both the Compilation Verifier and the Function Validator, the FSR framework constructs a new prompt for each candidate. If a candidate fails at the Function Validator specifically due to an output mismatch with the reference output, the corresponding Refined Prompt is structured as follows:

\begin{tcolorbox}[title=\textbf{Prompt},colback=SeaGreen!10!CornflowerBlue!10,colframe=RoyalPurple!55!Aquamarine!100!]
The result is not the same with the reference output. Modify the code.

The output should be the content of whole .cu file containing ONE kernel function.

Do not modify the test part.
\end{tcolorbox}

\section{Prompt of Tasks in Benchmarks}
\label{sec:prompt}

Table~\ref{tab:task_description} presents a high-level description of the target functionality or computational objective that each task is expected to achieve in the evaluation benchmarks. These descriptions are incorporated into the initial prompts to explicitly specify the intended task for each corresponding evaluation scenario.

\begin{table*}[!ht]
    \centering
    \begin{tabularx}{\textwidth}{ccX}
    \toprule
    \multicolumn{1}{c}{\textbf{Task ID}} & \multicolumn{1}{c}{\textbf{Task}} & \multicolumn{1}{c}{\textbf{Prompt}}  \\
       \midrule
1 & Sigmoid & Implement a CUDA program for sigmoid activation function: $\text{sigmoid}(x) = 1 / (1 + \exp(-x))$. Input shape: (batch\_size, dim); Output: same shape as input. \\ \midrule
2 & Matrix Multiplication & Write a program that multiplies two matrices of 32-bit floating point numbers on a GPU. Given matrix $A$ of dimensions $M \times K$ and matrix $B$ of dimensions $K \times N$, compute the product matrix $C = A \times B$, which will have dimensions $M \times N$. \\ \midrule
3 & Max Pooling 3D & Implement a CUDA program for 3D max pooling function that selects the maximum value within a defined region (a window) of a feature map. Input shape: (batch\_size, channels, dim1, dim2, dim3); Output: with 3D max pooling applied. \\ \midrule
4 & LayerNorm & Implement a GPU program that performs Layer Normalization (LayerNorm) operation, which normalizes across the features for each individual data sample in a layer. Input of shape (batch\_size, features, dim1, dim2); Output with Layer Normalization applied, same shape as input. \\ \midrule
5 & 2D Convolution & Write a program that performs a 2D convolution operation on the GPU. Given an input matrix and a kernel (filter), compute the convolved output. The convolution should be performed with a ``valid'' boundary condition, meaning the kernel is only applied where it fully overlaps with the input. The input consists of: (1) input: A 2D matrix of 32-bit floating-point numbers, represented as a 1D array in row-major order. (2) kernel: A 2D kernel (filter) of 32-bit floating-point numbers, also represented as a 1D array in row-major order. The output should be written to the output matrix (also a 1D array in row-major order). The output matrix will have dimensions: output\_rows = input\_rows - kernel\_rows + 1, output\_cols = input\_cols - kernel\_cols + 1. The convolution operation is defined as: $output[i][j] = \sum_{m=0}^{kernel\_rows-1} \sum_{n=0}^{kernel\_cols-1} input[i+m][j+n] * kernel[m][n]$. \\ \midrule
6 & Multi-Head Self-Attention & Implement a CUDA program for multi-head self-attention. Given three input matrices $Q$ (queries), $K$ (keys), and $V$ (values) of size $N \times d_{\text{model}}$, compute: $\text{MultiHead}(Q,K,V) = \text{Concat}(\text{head}_1,\ldots,\text{head}_h)$, where each head computes: $\text{head}_i = \text{softmax}\left(\frac{Q_iK_i^T}{\sqrt{d_k}}\right)V_i$ with $d_k = d_{\text{model}}/h$ and $Q_i$, $K_i$, $V_i$ being the $i$-th head's partition of the input matrices. \\ \midrule
7 & Mean Square Error & Implement a CUDA program to calculate the Mean Squared Error (MSE) between predicted values and target values. Given two arrays of equal length, predictions and targets, compute: $\text{MSE}=\frac{1}{N}\sum_{i=1}^{N}(predictions_{i}-targets_{i})^2$ where $N$ is the number of elements in each array. Input: predictions, targets; Output: MSE. \\ \midrule
\end{tabularx}
\end{table*}

\begin{table*}[!ht]
    \centering
    \begin{tabularx}{\textwidth}{ccX}
    \midrule
8 & Matrix Transpose & Write a program that transposes a matrix of 32-bit floating point numbers on a GPU. The transpose of a matrix switches its rows and columns. Given a matrix $A$ of dimensions rows $\times$ cols, the transpose $A^T$ will have dimensions cols $\times$ rows. All matrices are stored in row-major format.  \\ \midrule
9 & Reverse Array & Implement a program that reverses an array of 32-bit floating point numbers in-place. The program should perform an in-place reversal of input.  \\ \midrule
10 & ReLU Activation Fuction & Implement a program that performs the Rectified Linear Unit (ReLU) activation function on a vector of 32-bit floating point numbers. The ReLU function sets all negative values to zero and leaves positive values unchanged: $\text{ReLU}(x)=\max(0,x)$. \\ \midrule
11 & Top-K Selection & Implement a GPU program that, given a 1D array input of 32-bit floating point numbers of length $N$, selects the $k$ largest elements and writes them in descending order to the output array of length $k$. \\ \midrule
12 & Sorting & Write a CUDA program that sorts an array of 32-bit floating-point numbers in ascending order using the bubble sort algorithm. Do not use other algorithms. \\ \midrule
13 & Matrix Copy & Implement a program that copies an $N \times N$ matrix of 32-bit floating point numbers from input array $A$ to output array $B$ on the GPU. The program should perform a direct element-wise copy so that $B_{i,j} = A_{i,j}$ for all valid indices. \\ \midrule
14 & Reduction & Write a CUDA program that performs parallel reduction on an array of 32-bit floating point numbers to compute their sum. The program should take an input array and produce a single output value containing the sum of all elements. \\ \midrule
15 & Dot Product & Implement a CUDA program that computes the dot product of two vectors containing 32-bit floating point numbers. The dot product is the sum of the products of the corresponding elements of two vectors. Mathematically, the dot product of two vectors $A$ and $B$ of length $n$ is defined as: $A \cdot B = \sum_{i=0}^{n-1} A_i \cdot B_i$. \\ \midrule
16 & Prefix Sum & Write a CUDA program that computes the prefix sum (cumulative sum) of an array of 32-bit floating point numbers. For an input array $[a, b, c, d, \ldots]$, the prefix sum is $[a, a+b, a+b+c, a+b+c+d, \ldots]$. \\ \midrule
17 & Categorical Cross-Entropy Loss & Implement a CUDA program to calculate the categorical cross-entropy loss for a batch of predictions. Given a matrix of predicted logits $Z$ of size $N \times C$ and a vector of true class labels true\_labels of size $N$, compute the average cross-entropy loss over the batch. The loss for a single sample $j$ with logits $z_j = [z_{j1}, \ldots, z_{jC}]$ and true label $y_j$ is calculated using the numerically stable formula: $\text{Loss}_j = \log\left(\sum_{k=1}^{C} e^{z_{jk}}\right) - z_{j, y_j}$. The final output stored in the loss variable should be the average loss over the $N$ samples: $L = \frac{1}{N} \sum_{j=1}^{N} \text{Loss}_j$. Input: logits, true\_labels, $N$ (number of samples), and $C$ (number of classes). Output: loss (a pointer to a single float). \\ \midrule
18 & Monte Carlo Integration & Implement Monte Carlo integration on a GPU. Given a set of function values $y_i=f(x_i)$ sampled at random points uniformly distributed in the interval $[a, b]$, estimate the definite integral: $\int_{a}^{b}f(x)dx\approx (b-a)\cdot\frac{1}{n}\sum_{i=1}^{N}y_i$. The Monte Carlo method approximates the integral by computing the average of the function values and multiplying by the interval width. \\ \midrule
\end{tabularx}
\end{table*}

\begin{table*}[!ht]
    \centering
    \begin{tabularx}{\textwidth}{ccX}
    \midrule
19 & Histogramming & Write a GPU program that computes the histogram of an array of 32-bit integers. The histogram should count the number of occurrences of each integer value in the range [0, num\_bins). You are given an input array input of length $N$ and the number of bins num\_bins. The result should be an array of integers of length num\_bins, where each element represents the count of occurrences of its corresponding index in the input array. \\ \midrule
20 & Ordinary Least Squares Regression & Solve the Ordinary Least Squares (OLS) regression problem on a GPU. Given a feature matrix $X$ of size $n\_samples \times n\_features$ and a target vector $y$ of size $n\_samples$, compute the coefficient vector $\beta$ that minimizes the sum of squared residuals: $\min_{\beta} \|X\beta - y\|^2$. The closed-form solution to OLS is: $\beta = (X^TX)^{-1}X^Ty$. \\ \bottomrule
\end{tabularx}
\caption{High-level functional descriptions of evaluation tasks for initial prompt  construction.}
\label{tab:task_description}
\end{table*}

\end{document}